\documentclass[letterpaper]{article} 
\usepackage[submission]{aaai23}  
\usepackage{times}  
\usepackage{helvet}  
\usepackage{courier}  
\usepackage[hyphens]{url}  
\usepackage{graphicx} 
\usepackage{natbib}  
\usepackage{caption} 
\usepackage{algorithm}
\usepackage{algorithmic}

\usepackage{newfloat}
\usepackage{listings}
\DeclareCaptionStyle{ruled}{labelfont=normalfont,labelsep=colon,strut=off} 
\floatstyle{ruled}
\newfloat{listing}{tb}{lst}{}
\floatname{listing}{Listing}
\title{STSC-SNN: Spatio-Temporal Synaptic Connection with Temporal Convolution and Attention for Spiking Neural Networks}
\author{
    Chengting Yu \textsuperscript{\rm 1,2},
    Zheming Gu \textsuperscript{\rm 1},
    Da Li \textsuperscript{\rm 1},
    Gaoang Wang \textsuperscript{\rm 2},
    Aili Wang \textsuperscript{\rm 1,2,*},
    Erping Li \textsuperscript{\rm 1,2}
}
\affiliations{
    \textsuperscript{\rm 1} College of Information Science and Electronic Engineering, Zhejiang University, Hangzhou, China \\ 
    \textsuperscript{\rm 2} ZJU-UIUC Institute, Zhejiang University, Haining, China \\
    chengting.21@intl.zju.edu.cn,
    ailiwang@intl.zju.edu.cn
    
}

\begin{document}

\maketitle

\begin{abstract}
Spiking Neural Networks (SNNs), as one of the algorithmic models in neuromorphic computing, have gained a great deal of research attention owing to temporal information processing capability, low power consumption, and high biological plausibility.
The potential to efficiently extract spatio-temporal features makes it suitable for processing the event streams.
However, existing synaptic structures in SNNs are almost full-connections or spatial 2D convolution, neither of which can extract temporal dependencies adequately.
In this work, we take inspiration from biological synapses and propose a spatio-temporal synaptic connection SNN (STSC-SNN) model, to enhance the spatio-temporal receptive fields of synaptic connections, thereby establishing temporal dependencies across layers.
Concretely, we incorporate temporal convolution and attention mechanisms to implement synaptic filtering and gating functions.
We show that endowing synaptic models with temporal dependencies can improve the performance of SNNs on classification tasks. In addition, we investigate the impact of performance vias varied spatial-temporal receptive fields and reevaluate the temporal modules in SNNs.
Our approach is tested on neuromorphic datasets, including DVS128 Gesture (gesture recognition), N-MNIST, CIFAR10-DVS (image classification), and SHD (speech digit recognition). The results show that the proposed model outperforms the state-of-the-art accuracy on nearly all datasets.
\end{abstract}

\section{Introduction}
Spiking neural networks (SNNs) are regarded as the third generation of neural networks \cite{maass_networks_1997}, with the purpose of addressing the fundamental mysteries of intelligence and the brain by emulating biological neurons and incorporating more biological mechanisms \cite{roy_towards_2019}. The two fundamental components of SNNs are spiking neurons and synapses, which create a hierarchical structure (layers) and subsequently construct a network. 
SNNs have attracted a significant deal of academic interest in recent years due to their prospective properties, such as the ability to process temporal information, low power consumption \cite{roy_towards_2019}, and biological interpretability \cite{gerstner_neuronal_2014}.
Currently, SNNs are capable of processing event stream data with low latency and low power \cite{pei_towards_2019,gallego_event-based_2020}. However, there is still a performance gap between SNNs and traditional Artificial Neural Networks (ANNs). Recent SNN training techniques based on surrogate gradients and back-propagation have significantly enhanced the performance of SNNs \cite{wu_spatio-temporal_2018,fang_incorporating_2021}, while also promoting the further integration of ANNs' modules into SNNs
\cite{zheng_going_2020,hu_spiking_2021,yao_temporal-wise_2021},
greatly accelerating the development of SNNs. 
However, it remains challenging to connect these computational techniques with the biological properties of SNNs.

\begin{figure}[t]
\centering
\includegraphics[width=0.9\columnwidth]{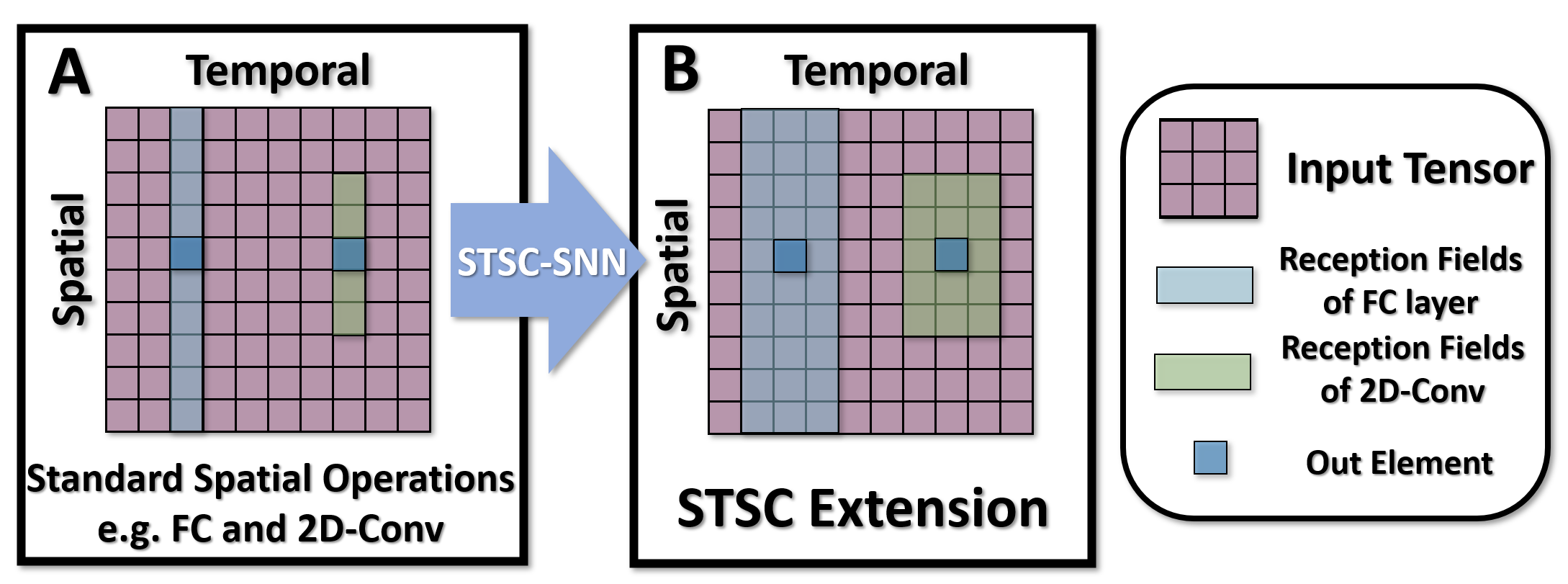} 
\caption{Illustration of Receptive Fields in Synaptic Connections. (a) The receptive fields of typical spatial operations used in SNNs, e.g., fully-connected layers (full) and 2D convolutional layers (sparse); (b) The STSC modules proposed to extend spatial operations with spatio-temporal receptive fields. 
}.
\label{fig1}
\end{figure}

Due to the time-dependent correlation of neuron dynamics, it is believed that SNNs naturally process information in both temporal and spatial dimensions \cite{roy_towards_2019}. Further researches are necessary to harness the spatio-temporal information processing capabilities of SNNs. Combining ANNs’ modules has significantly increased the performance of SNNs in several research studies. In terms of spatial information processing, CSNN \cite{xu_csnn_2018} was the first to validate the application of convolution structure on SNNs, followed by the proposal of NeuNorm to improve SNNs’ usage of convolution through auxiliary neurons \cite{wu_direct_2019}. In the time dimension, \cite{zheng_going_2020} implements the time-dependent batch normalization (tdBN) module to tackle the issue of gradient vanishing and threshold balancing, and \cite{yao_temporal-wise_2021} uses the Squeeze-and-Excitation (SE) block \cite{hu_squeeze-and-excitation_2018} to realize the attention distribution of the temporal dimension in order to improve the temporal feature extraction. Notably, \cite{zhu_tcja-snn_2022} proposes Temporal-Channel Joint Attention (TCJA) to concurrently process input in both temporal and spatial dimensions, which is a significant effort for SNNs' spatio-temporal feature extraction. These studies effectively improve the performance of SNNs by transplanting established ANNs' modules and methodologies. However, applying these computational modules to SNNs from the standpoint of deep learning dilutes the fundamental biological interpretability, bringing SNNs closer to a mix of existing concepts in machine learning, such as recurrent neural networks (RNNs), binary neural networks (BNNs), and quantization networks.

From a biological standpoint, some works focus on the synapse models, investigating the potential of SNNs in respect of connection modes and information transmission. \cite{shrestha_slayer_2018,fang_exploiting_2020,yu_map-snn_2022} integrate impulse response models with synaptic dynamics, hence enhancing the temporal information representation of SNNs; \cite{cheng_lisnn_2020} implements intra-layer lateral inhibitory connections to improve the noise tolerance of SNNs; from the standpoint of synaptic plasticity, \cite{bellec_solution_2020,zhang_spike-train_2019} introduce bio-plausible training algorithms as an alternative to back-propagation (BP), allowing for lower-power training. Experiments revealed that the synaptic models of SNNs have a great deal of space for modification and refinement in order to handle spatio-temporal data better \cite{fang_exploiting_2020}. We propose a Spatio-Temporal Synaptic Connection (STSC) module for this reason.

Based on the notion of spatio-temporal receptive fields, the structural features of dendritic branches \cite{letellier_differential_2019} and feedforward lateral inhibition \cite{luo_architectures_nodate} motivate this study. By merging the ANNs’ computation modules (temporal convolutions and attention mechanisms) with SNNs, we propose the
STSC module,
consisting of Temporal Response Filter (TRF) module and Feedforward Lateral Inhibition (FLI) module. As shown in Fig. \ref{fig1}, the STSC can be attached to spatial operations to expand the spatio-temporal receptive fields of synaptic connections, hence facilitating the extraction of spatio-temporal features. 
The main contributions of this work are summarized as follows:
\begin{itemize}
    \item We propose STSC-SNN to implement synaptic connections with extra temporal dependencies and enhance the SNNs' capacity to handle temporal information. To the best of our knowledge, this study is the first to propose the idea of synaptic connections with spatio-temporal receptive fields in SNNs and to investigate the influence of synaptic temporal dependencies in SNNs.
    \item Inspired by biological synapses, we propose two plug-and-play blocks: Temporal Response Filter (TRF) and Feedforward Lateral Inhibition (FLI), which perform temporal convolution and attention operations and can be simply implemented into deep learning frameworks for performance improvements.
    \item On neuromorphic datasets, DVS128 Gesture, SHD, N-MNIST, CIFAR10-DVS, we have produced positive results. Specifically, we acquire 92.36\% accuracy on SHD with a simple fully-connected structure, which is a great improvement above the 91.08\% results obtained with recurrent structure and reaches performance comparable to ANNs.
\end{itemize}

\section{Related Work}

\subsection{Learning algorithms for SNNs}
In recent years, many works have explored the learning algorithms of SNNs, which can be generally categorized as biologically inspired approaches \cite{diehl_unsupervised_2015,bellec_solution_2020,zhang_spike-train_2019}, ANN-to-SNN conversion methods \cite{orchard_converting_2015,sengupta_going_2019,han_rmp-snn_2020}, and surrogate-based direct training methods \cite{wu_spatio-temporal_2018,neftci_surrogate_2019,fang_incorporating_2021}.
Direct training methods utilize surrogate gradients to tackle the issue of non-differentiable spike activity \cite{wu_spatio-temporal_2018}, allowing error back-propagation (BP) through time to interface the gradient descent directly on SNNs for training. Those BP-based methods show strong potential to achieve high accuracy in a few timesteps by making full use of spatio-temporal information\cite{wu_direct_2019,fang_incorporating_2021}. 
However, more research is required to determine how to better extract spatio-temporal features for enhanced processing of spatio-temporal data; this is what we want to contribute.

\subsection{Attention Modules in SNNs}
The attention mechanism distributes attention preferentially to the most informative input components, which could be interpreted as the sensitivity of various inputs. The SE block \cite{hu_squeeze-and-excitation_2018} offers an efficient attention approach to improve representations in ANNs. \cite{xie_efficient_2017,kundu_spike-thrift_2021} introduced spatial-wise attention in SNNs; then, TA-SNN \cite{yao_temporal-wise_2021} developed a temporal-wise attention mechanism in SNNs by assigning attention factors to each input frame; more subsequently, TCJA \cite{zhu_tcja-snn_2022} added a channel-wise attention module
and proposed temporal-channel joint attention. 
These studies demonstrate the usefulness of attention mechanisms in SNNs by achieving state-of-the-art results on various datasets. 
Moreover, based on these investigations, it is desirable to study other correlations between the attention mechanism and the biological nature of SNNs, which is the objective of our research. We employ the attention module as a feedforward lateral inhibitory connection \cite{luo_architectures_nodate}, which develops a gating mechanism for the synapse model, and enables nonlinear computation by the synapse.

\subsection{Synaptic Models in SNNs}
As one of the fundamental components of SNN, the synaptic model has drawn the interest of several researchers. \cite{shrestha_slayer_2018,fang_exploiting_2020,yu_map-snn_2022} established temporal relationships between response post-synaptic currents and input pre-synaptic spikes, therefore improving temporal expressiveness. Those temporal relationships are the extension of fully-connected synapses
which are based on the assumption that there is only one connection between two neurons.
Nevertheless, synaptic connections are often complex, and there are typically many paths connecting the axons and dendrites of neurons \cite{luo_architectures_nodate,letellier_differential_2019}. 
We apply temporal convolution to describe the more sophisticated impulse response model and generate time-dependent post-synaptic currents, taking into consideration biological features and computational simplicity.

\begin{figure}[t]
\centering
\includegraphics[width=0.9\columnwidth]{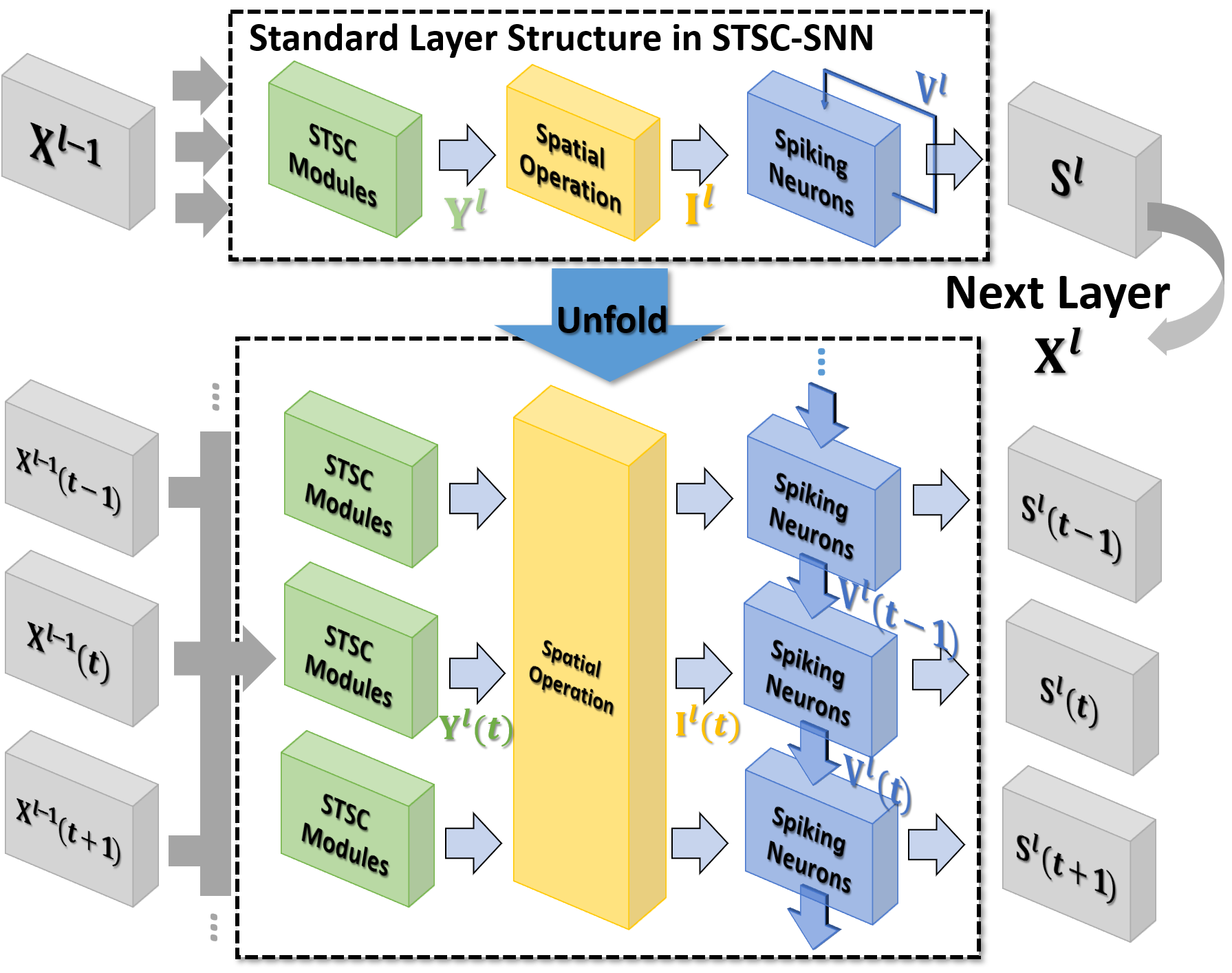} 
\caption{The standard layer inserted with the STSC module and its unfolded formulation. Note that all parameters are shared at all timesteps. STSC modules are set before spatial operations to process the latest temporal information.
}.
\label{fig2}
\end{figure}
\section{Approach}
\subsection{Frame-based Representation}
Event steam consists of both a spatial and a temporal dimension, with the spatial dimension expandable to higher dimensions depending on the data type.
The spatial dimension of event streams based on sound data is typically one-dimensional, corresponding to different frequency channels; whereas the spatial dimension of event streams based on image data is typically three-dimensional, consisting of coordinates representing spatial positions and polarities representing brightness changes.
The binary spike pattern is represented by the tensor $E\in B^{T'\times S}$, where $T'$ represents the original resolution in the temporal dimension, and $S$ represents the resolution in the spatial dimension.
For a frame with a time span of $\Delta t$, the events in the time interval $t'\in [(t-1)\times \Delta t,t\times \Delta t)$ can be mapped to the network input $\mathbf{X}^0$ at time $t$ by
\begin{equation}
    x^0(t)=q(\{E(t')|t'\in [(t-1)\times \Delta t,t\times \Delta t)\} )
\end{equation}
where $t\in\{1,2,...,T\}$ is timesteps, and the aggregation function $q(\cdot)$ could be chosen as non-polarity aggregation \cite{massa_efficient_2020}, accumulate aggregation \cite{deng_rethinking_2020}, AND aggregation \cite{he_comparing_2020} etc. Here, we choose to accumulate all event streams inside a frame (see Fig. \ref{fig7}).

\subsection{Spiking Neurons in SNNs}
The Leaky-Integrate-and-Fire (LIF) model was introduced as an extremely simplified model of biological neurons \cite{dayan_theoretical_2005}, which has the essential qualities of potential integrating, leaking, and spike firing. 
The LIF model is used extensively in SNNs and neuromorphic engineering because of its ability to recreate essential neural functions at a minimal cost of computation.
The LIF model is defined in the differential form, as
\begin{equation}
    \tau\frac{dv(t)}{dt} = - v(t)  + I(t)
\end{equation}
where $v(t)$ is the membrane potential of the neuron at time $t$, $I(t)$ is the integrated current input from the pre-synaptic neuron at time $t$, and $\tau$ is the time constant that governs the pace of potential change.
Solving the differential equation directly will incur additional costs. STBP \cite{wu_spatio-temporal_2018} employs a simplified iterative representation and implements the LIF model on the Pytorch framework \cite{paszke_pytorch_2019}, which supports the integration of SNNs and standard ANNs' modules and significantly speeds the construction of BP-based SNNs and training techniques. The explicit iterative LIF is expressed as
\begin{equation}
\mathbf{V}^{l}\left( t \right) = \left( {1 - \frac{1}{\mathbf{\tau}}} \right) \times \mathbf{V}^{l}\left( {t - 1} \right) \times \left( {1 - \mathbf{S}^{l}\left( { t - 1} \right)} \right) + \mathbf{I}^{l}\left( t \right)    
\end{equation}
\begin{equation}
\mathbf{S}^{l}\left( t\right) = \mathbf{\Theta}\left( \mathbf{V}^{l}\left( t \right) - V_{th} \right)
\end{equation}
where $l$ and $t$ are indices of layer and time, $\tau$ is the time constant, $\mathbf{V}$ is the membrane potential, $V_{th}$ is the threshold constant, $\mathbf{S}$ is the binary tensor of spikes, $\mathbf{I}$ is the input from the preceding layer, and ${\Theta} (\cdot)$ is the Heaviside step function.
Noting that the firing process, ${\Theta} (\cdot)$, is not differentiable, surrogate methods are often utilized in SNNs' direct training to achieve error propagation by creating various pseudo-derivatives for ${\Theta} (\cdot)$ \cite{neftci_surrogate_2019}. 
This work leverages arc tangent (ATan) as the pseudo-derivative of ${\Theta} (\cdot)$, which is well supported in the SpikingJelly framework \cite{SpikingJelly}. 

\begin{figure}[t]
\centering
\includegraphics[width=0.9\columnwidth]{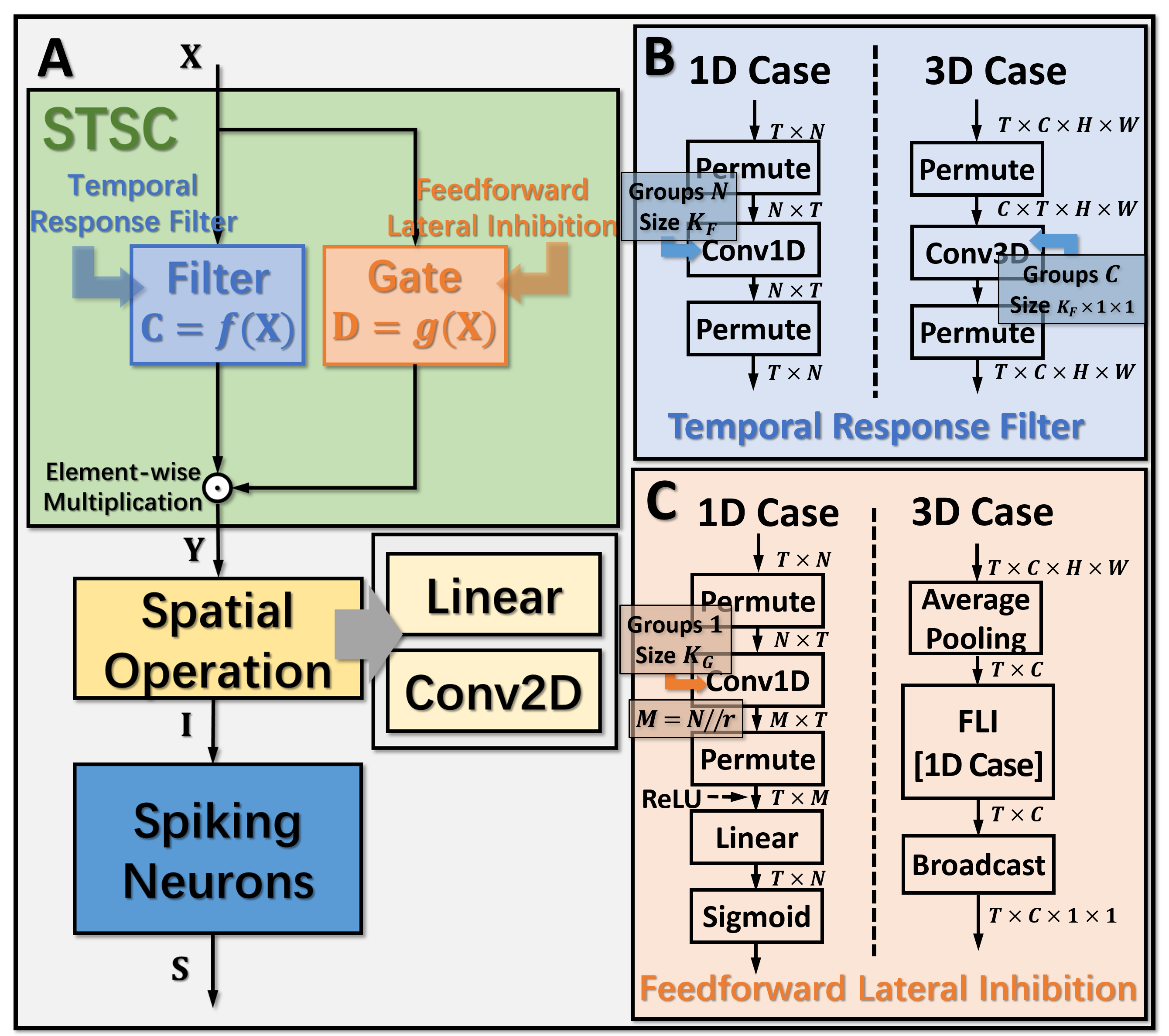} 
\caption{Operation Details of STSC modules. (a) Connection Implementation between TRF, FLI, and Spatial Operations, where $\odot$ denotes broadcast element-wise multiplication; (b) Tensor Computations in TRF; (c) Tensor Computations in FLI. 
}.
\label{fig3}
\end{figure}

\begin{figure}[t]
\centering
\includegraphics[width=0.9\columnwidth]{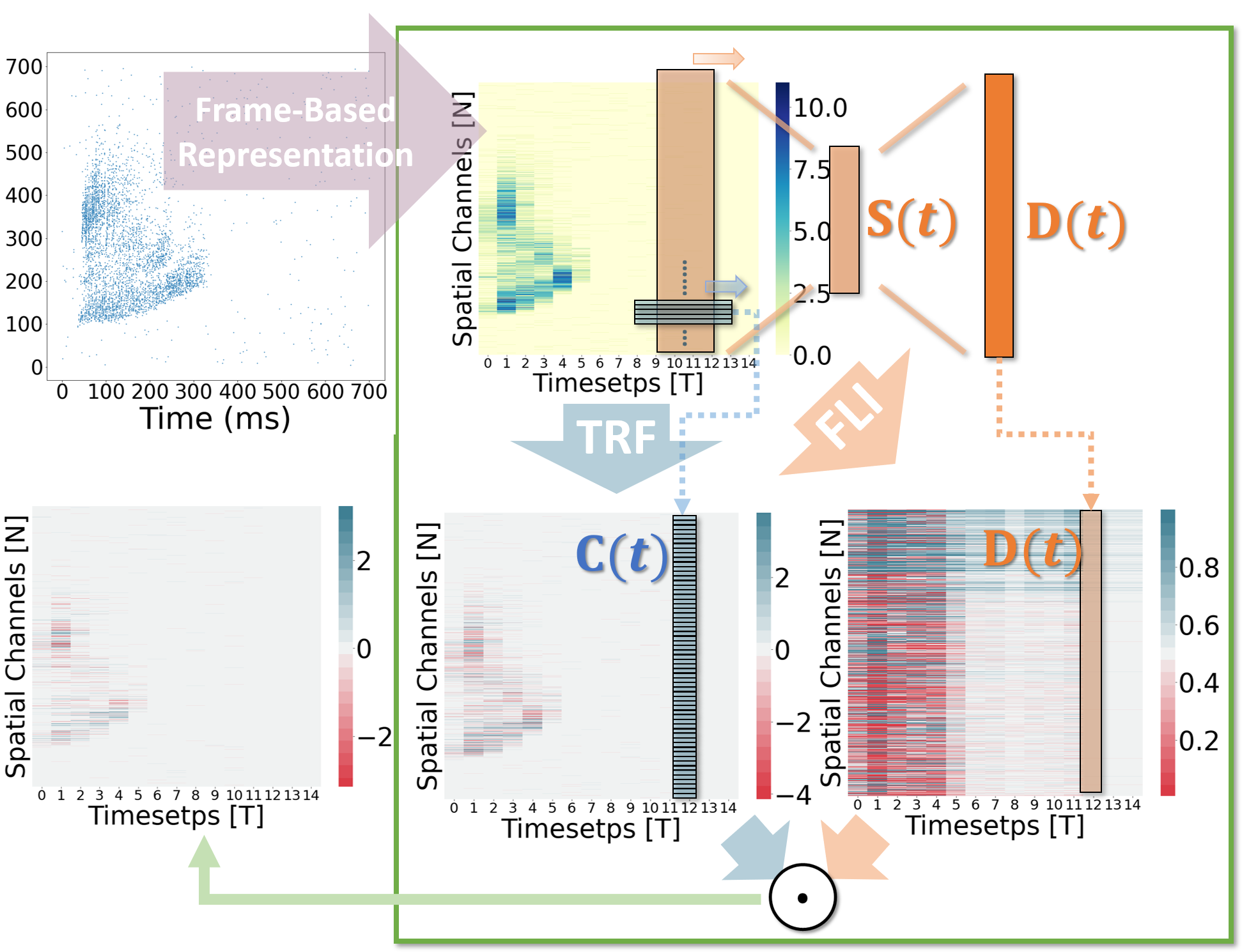} 
\caption{Computation Visualization. The input is an audio sample from the SHD dataset.
}.
\label{fig7}
\end{figure}

\subsection{Spatio-Temporal Receptive Fields in SNNs}
The receptive field is often used to comprehend convolution procedures. In the process of convolution, the receptive fields describe the range of the nearby input for identifying an output element, i.e., how much spatial neighboring position it can perceive.
For static pictures, the receptive field could explain the projection range of the convolution operations and aid in the comprehension of the spatial feature extraction procedure. 
Similarly, the concept of receptive fields could be applied to event streams (or dynamic images) with an additional temporal dimension.
This work leverages the concept of Spatio-Temporal Receptive Fields to aid comprehension of SNNs' spatio-temporal feature extraction procedure. 
As shown in Fig. \ref{fig1}a,  typical synaptic connections employ 2D convolution, pooling, full-connections and other inter-layer computations to process information in the spatial dimension, which we refer to as spatial operations, and their receptive fields are restricted to the spatial dimension.
To strengthen the spatio-temporal information processing capabilities of SNNs, it is essential to expand the receptive fields of these spatial operations into the temporal dimension.

\subsection{Spatio-Temporal Synaptic Connection for SNNs}
In general, the processing of temporal information in SNNs is attributed to spiking neurons, since their dynamic model has a natural dependence on the temporal dimension; however, the level of this dependence is primarily reliant on the degree of neural complexity, while the LIF neurons only support very weak temporal linkages.
Not just in neurons, but also in biological synapses, a great deal of the processing of latent temporal characteristics occurs \cite{luo_architectures_nodate,letellier_differential_2019}.
This work focus on using temporal dimension operations in SNNs to broaden the spatio-temporal receptive fields of synapses, to enhance the spatio-temporal feature extractions of SNNs.
Temporal operations are calculations connected to the time dimension, and they are contained in a pluggable module, referred to as the Spatio-Temporal Synaptic Connection (STSC). The STSC module is designed to be placed before spatial operations in order to aggregate  temporal information and enlarge the spatio-temporal receptive fields while maintaining the original spatial operations (see Fig. \ref{fig2}).
The STSC module consists of two modules: Temporal Response Filter and Feedforward Lateral Inhibition, which carry the filtering and gating mechanisms of the synaptic model, respectively (see Fig\ref{fig3}). The two modules receive $\mathbf{X}$ as input tensor and conduct the operations $\mathbf{C}=f(\mathbf{X})$ and $\mathbf{D}=g(\mathbf{X})$, followed by element-wise product to produce output $\mathbf{Y}=\mathbf{C}\odot \mathbf{D}$ (see Fig. \ref{fig3}a \& Fig. \ref{fig7}).
\subsubsection{Temporal Response Filter}
The synapses in biological neural networks are intricate. The complexity of synapses is not only reflected in the non-topological spatial relationship of synapses (how neurons connect with one another) but also in the complicated temporal dependency of spike transmission (how neurons communicate with one another) \cite{letellier_differential_2019}. 
This work proposes the Temporal Response Filter (TRF) to establish the linear response of spikes over time by employing convolution in the time dimension, in order to expand the temporal receptive field in the most direct way. TRF offers a filtering path for STSC with temporal convolutions (see Fig. \ref{fig3}). 
Fig. \ref{fig3}b depicts the specific implementation of TRF. In detail, as for the 2D spatiotemporal tensor in the fully-connected structure, it performs temporal depth-wise 1D convolution independently on each spatial channel and generates an output tensor of the same size. To ensure that all spatial elements inside a channel have the same temporal response, for the 4D spatiotemporal tensor in the convolutional structure, temporal depth-wise 3D convolution is performed on each channel with kernel size of $K_{G} \times 1 \times 1$.
Mathematically, the filter operation is denoted as $f(\cdot)$, and it performs $\mathbf{C}=f(\mathbf{X})$ with input $\mathbf{X}$ and output $\mathbf{C}$ having the same size as $R^{T\times N}$ or $R^{T\times C\times H\times W}$.
Depending on the spatial dimension (1D or 3D) of the input $\mathbf{X}$, the following calculation formulae apply:\\
As for the 1D case,
\begin{equation}
    \mathbf{C}\left( t,n \right) = {\sum\limits_{t_f = - \frac{K_F - 1}{2}}^{\frac{K_F - 1}{2}}{\mathbf{W}^{F}_{t_f,n} \times \mathbf{X}\left( t - t_f,n \right)}}
\end{equation}
As for the 3D case,
\begin{equation}
\mathbf{C}\left( t,c,h,w \right) = {\sum\limits_{t_f = - \frac{K_F - 1}{2}}^{\frac{K_F - 1}{2}}{\mathbf{W}^F_{t_f,c} \times \mathbf{X}\left( t - t_f,c,h,w \right)}} 
\end{equation}
where $n$,$c$,$h$, and $w$ are spatial location indices and $t$ is a time index. $K_F$ denotes the kernel size of the temporal convolution, which is equal to the temporal receptive fields of TRF. The padding of the convolution is set to $\frac{K_F-1}{2}$ for maintaining the same size.
\subsubsection{Feedforward Lateral Inhibition}
The mechanisms of feedforward lateral inhibition mechanisms exist in biological neural networks \cite{luo_architectures_nodate}, which construct a lateral route to suppress feedforward input. We notice that the function of this structure closely resembles that of the attention module; therefore, we refer to the attention blocks \cite{hu_squeeze-and-excitation_2018,yao_temporal-wise_2021,zhu_tcja-snn_2022}, and propose the FLI module to replicate the gating mechanism in synaptic connections.
The module details are shown in Fig. \ref{fig3}c. 
Regarding the 2D spatiotemporal tensor in the fully-connected structure, temporal-wise 1D convolution is utilized first to extract temporal features, followed by linear combination through sigmoid to acquire the gating coefficients (see Fig. \ref{fig3}c).
As for the 4D spatiotemporal tensor in the convolutional structure, spatial-wise average pooling is first conducted to obtain the channel-wise spatial sparsity of spikes; then, the 1D case FLI is performed. Finally, channel-wise gating factors are computed and transmitted to each channel's spatial locations.
Mathematically, gating is denoted as $g(\cdot)$, and $\mathbf{X}$ is the input tensor of size $R^{T\times N}$ or $R^{T\times C\times H\times W}$, $\mathbf{D}=f(\mathbf{X})$ is the output gating factors with values in the range $(0,1)$ that have the same shape with $\mathbf{X}$. Depending on the spatial dimension (1D or 3D) of the input $\mathbf{X}$, the following calculation equations apply:\\
As for the 1D case:
\begin{equation}
\mathbf{S}\left( {t,m} \right) = {\sum\limits_{t_g = - \frac{K_G - 1}{2}}^{\frac{K_G - 1}{2}}{\sum\limits_{n=1}^{N}{\mathbf{W}^{G1}_{t_g,n,m} \times \mathbf{X}\left( {t - t_g,n} \right)}}}    
\end{equation}
\begin{equation}
\mathbf{D}\left( {t,n} \right) = \mathbf{Sigmoid}\left( {\sum\limits_{m = 1}^{M}{\mathbf{W}^{G2}_{m,n} \times \mathbf{ReLU}\left( {\mathbf{S}\left( {t,m} \right)} \right)}} \right)    
\end{equation}
As for the 3D case:
\begin{equation}
\hat{\mathbf{X}}\left( t,c \right) = \frac{1}{H \times W}{\sum\limits_{h = 1}^{H}{\sum\limits_{w = 1}^{W}{\mathbf{X}\left( t,c,h,w \right)}}}
\end{equation}
\begin{equation}
\hat{\mathbf{D}} = g_{\mathbf{1D}}\left( \hat{\mathbf{X}} \right)
\end{equation}
\begin{equation}
\mathbf{D}\left( {t,c,h,w} \right) = \mathbf{~}\hat{\mathbf{D}}\left( {t,c} \right)\mathbf{~}\mathbf{~}\mathbf{~}\mathbf{~}\mathbf{f}\mathbf{o}\mathbf{r}\mathbf{~}\forall h \in H,\forall w \in W
\end{equation}
where $n$,$c$,$h$ and $w$ are spatial location indices, $t$ is a time index, $m$ is the index of the intermediate feature tensor $S$ with spatial dimension $M$. $M$ is determined by the spatial sizes $N$ with reduction ratio $r$, as $\frac{N}{r}$. $K_G$ denotes the kernel size of the convolution, which is equivalent to the receptive fields of FLI. The padding of the convolution is set to $\frac{K_G-1}{2}$ for maintaining the same size.

\subsection{Training Framework}
Denote the simulating timesteps as $T$, size of output layers as $L_{out}$ and classes number as $C$, we utilize the voting strategy \cite{wu_direct_2019} to decode the network output $O\in B^{T\times L_{out}}$ with the constant voting matrix $M\in R^{C\times L_{out}}$. The loss function is defined by the mean squared error (MSE), as
\begin{equation}
    L = \left| \left| {\mathbf{y}_{i} - \frac{1}{T}\sum M_{i,n}O\left( {t,n} \right)} \right| \right|^{2}
\label{eq:loss}
\end{equation}
where $\mathbf{y}$ is the one-hot target, with $\mathbf{y}_l=1$ for target class $l$, and $\mathbf{y}_i=0$ for $i\neq l$. The predicted label $l_p$ is then given by $l_p=argmax_i  \frac{1}{T}\sum M_{i,n}O\left( {t,n} \right)$ for evaluation. In the experiment, we adopted the simplest voting strategy and obtained $\sum M_{i,n}O\left( {t,n} \right)$ through average pooling.

\begin{table}[t]
\centering
\resizebox{.95\columnwidth}{!}{
\begin{tabular}{|l|l|}
\hline
Datasets             & \begin{tabular}[c]{@{}l@{}}SHD \& CIFAR10-DVS \& \\ DVS128 Gesture \& N-MNIST\end{tabular} \\
\hline
Representation       & Frames with Accumulative Aggregation                                                                             \\
Learning   Algorithm & STBP \cite{wu_spatio-temporal_2018} \& BPTT                                                                         \\
Surrogate   Gradient & ATan \cite{fang_incorporating_2021}                                                                               \\
Loss Function        & Voting \cite{wu_direct_2019} \&   MSE   \\
Frameworks          & SpikingJelly  \& Pytorch  \\
\hline
\end{tabular}
}
\caption{Experimental Details. }
\label{tab:1}
\end{table}

\begin{table}[t]
\centering
\resizebox{.95\columnwidth}{!}{
\begin{tabular}{|c|c|c|c|c|}
\hline
Hyper Param.         & SHD  & N-MNIST & \begin{tabular}[c]{@{}c@{}}CIFAR10-\\ DVS\end{tabular}& \begin{tabular}[c]{@{}c@{}}DVS128 \\ Gesture\end{tabular}\\
\hline
Epoch                    & 200  & 300     & 1000        & 1000           \\
Batch Size               & 256  & 16      & 16          & 16             \\
Learning Rate            & 1e-4 & 1e-3    & 1e-3        & 1e-3           \\
$T$             & 15   & 10      & 10          & 20             \\
$\tau$            & 10   & 2       & 2           & 2              \\
$V_{th}$   & 0.3  & 1.0     & 1.0         & 1.0            \\
$K_F$ & 5    & 3       & 3     & 3              \\
$K_G$ & 3    & 3       & 3     & 5              \\
$r$        & 1    & 1       & 2           & 2     \\
\hline
\end{tabular}
}
\caption{Hyper-parameter Setting.}
\label{tab:2}
\end{table}

\begin{table}[t]
\centering
\resizebox{.95\columnwidth}{!}{
\begin{tabular}{|c|c|}
\hline
Dataset        & Network Structure                                                                                                                       \\ \hline
DVS128 Gesture & \begin{tabular}[c]{@{}c@{}}Input-128C3-MP2-128C3-MP2-128C3-\\ MP2-128C3-MP2-128C3-MP2-0.5DP-\\ 512FC-0.5DP-110FC-Voting-11\end{tabular} \\ \hline
CIFAR10-DVS    & \begin{tabular}[c]{@{}c@{}}Input-64C3-128C3-AP2-256C3-256C3-\\ AP2-512C3-512C3-AP2-512C3-512C3-\\ AP2-100FC-Voting-10\end{tabular}      \\ \hline
SHD            & Input-128FC-128FC-100FC-Voting-20                                                                                                       \\ \hline
N-MNIST        & \begin{tabular}[c]{@{}c@{}}Input-128C3-AP2-128C3-AP2-0.5DP-\\ 2048FC-0.5DP-100FC-Voting-10\end{tabular}                                 \\ \hline
\end{tabular}
}
\caption{Network Structure. 
The spiking neurons are added behind all $x$C$y$ and $n$FC. The STSC modules are inserted before all $x$C$y$ on DVS128 Gesture, CIFAR10-DVS, and N-MNIST; while inserted before all $n$FC on SHD.}
\label{tab:3}
\end{table}

\begin{table*}[t]
\centering
\resizebox{1.9\columnwidth}{!}{
\begin{tabular}{|l|cccccccc|}
\hline
    Method       & \multicolumn{2}{c}{SHD} & \multicolumn{2}{c}{N-MNIST} & \multicolumn{2}{c}{CIFAR10-DVS}                                      & \multicolumn{2}{c|}{DVS128 Gesture} \\
           &      T     &       Acc. (\%)      & T           & Acc. (\%)           & T  & Acc. (\%)                                                          & T    & Acc. (\%)               \\
\hline
NeuNorm  \cite{wu_direct_2019}  &      -     & -           & -           & 99.53         & -  & 60.5                                                            & -    & -                  \\
Rollout  \cite{kugele_efficient_2020}  &    -       & -           & 32          & 99.57         & 48 & 66.97                                                           & 240  & \begin{tabular}[c]{@{}c@{}}97.27\\   (10 classes)\end{tabular} \\
LISNN  \cite{cheng_lisnn_2020}    &     -      & -           & 20          & 99.45         & -  & -                                                               & -    & -                  \\
tdBN    \cite{zheng_going_2020}   &      -     & -           & -           & -             & 10 & 67.8                                                            & 40   & 96.87              \\
LIAF-Net \cite{wu_liaf-net_2021}  &           & -           & -           & -             & 10 & 70.40                                                           & 60   & 97.56              \\
PLIF    \cite{fang_incorporating_2021}   &      -     & -           & 10          & 99.61         & 20 & 74.80                                                           & 20   & 97.57              \\
LIF RSNN \cite{cramer_heidelberg_2020} &    2000      & 73.3         & -           & -            & - & -& -   & -             \\
Hetero. RSNN \cite{perez-nieves_neural_2021} &    -      & 83.5         & -           & 97.5             & - & - & -   & 82.9             \\
Adaptive SRNN    \cite{yin_accurate_2021}   &   250        & 90.4           & -           & -             & - & -& -   & -             \\
SEW-ResNet \cite{fang_deep_2021} &     -      & -           & -           & -             & 16 & 74.4                                                            & 16   & 97.92              \\
TA-SNN  \cite{yao_temporal-wise_2021}   & 15        & 91.08       & -           & -             & 10 & 72.00                                                           & 20   & 98.61              \\
TET    \cite{deng_temporal_2022}    &    -       & -           & -           & -             & 10 & \textbf{83.32}                                                           & -    & -                  \\
DSR    \cite{meng_training_2022}    &    -       & -           & -           & -             & 10 & 77.41                                                           & -    & -                  \\
TCJA    \cite{zhu_tcja-snn_2022}   &   -        & -           & -           & -             & 10 & \begin{tabular}[c]{@{}c@{}}80.7(MSE)\\   \textbf{83.3(TET)} \end{tabular} & 20   & \textbf{99.0}               \\

\hline
STSC (this work)      &      15     & \textbf{92.4}        & 10          & \textbf{99.64}         & 10 & \textbf{81.4(MSE)}                                                          & 20   & \textbf{98.96}        \\
\hline
\end{tabular}
}
\caption{Performance comparison between the proposed method and the state-of-the-art methods on different datasets.}
\label{tab:4}
\end{table*}

\section{Experiments}
\subsection{Experiment Setup}

\subsubsection{Datasets}
We evaluate the classification performance of STSC-SNN on a variety of neuromorphic datasets, including DVS128 Gesture \cite{amir_low_2017} (gesture recognition), N-MNIST \cite{orchard_converting_2015} and CIFAR10-DVS \cite{li_cifar10-dvs_2017} (image classification), and SHD \cite{cramer_heidelberg_2020} (speech digit recognition), all of which are event datasets but are generated using different methods. 
DVS128 Gesture is a gesture recognition dataset that uses DVS cameras to record actual human gestures.  The event-based image datasets, N-MNIST and CIFAR10-DVS, are converted from the static dataset by using DVS cameras to scan each sample. Spiking Heidelberg Digits (SHD) is a spike-based speech dataset consisting of English and German spoken digits transformed from the audio recordings using an artificial inner ear model.
\subsubsection{Learning}
Tab. \ref{tab:1} summarizes the experimental details of the SNNs training process. We use the SpikingJelly \cite{SpikingJelly} and Pytorch \cite{paszke_pytorch_2019} frameworks to develop and evaluate SNNs. We utilize the Adam optimizer \cite{kingma_adam_2017} to accelerate the training process.
Tab. \ref{tab:2} displays the respective hyper-parameters and Tab. \ref{tab:3} displays the network architectures for different datasets. All Conv2d layers are set as kernel size=3, stride=1, and padding=1, followed by batch normalization (BN) layers. The voting layers are implemented using average pooling for classification robustness \cite{fang_incorporating_2021}.

\subsection{Comparison with Existing SOTA Works}
Tab. \ref{tab:4} shows the performance comparison of the proposed methods (STSC-SNN with TRF and FLI) and other competing methods on neuromorphic datasets, N-MNIST, CIFAR10-DVS, DVS128 Gesture, and SHD. As shown in Tab. \ref{tab:4}, we achieve the highest accuracy on all datasets except CIFAR10-DVS. The SOTA results implemented in CIFAR10-DVS are based on the work of TET \cite{deng_temporal_2022}, which proposes a new loss function to enable the model to converge on a flatter local minimum with generalizability; TCJA \cite{zhu_tcja-snn_2022} also demonstrates its efficacy on CIFAR10-DVS. To preserve the consistency of this work, we continue to utilize MSE (Eq. (\ref{eq:loss})) as the loss function, and outperform the comparable result.
Notably, the experiments on SHD show that we have enhanced the vanilla SNN from 78.71\% to 92.36\% using STSC, which is the state-of-the-art result compared to the highest available result (91.08\% by TA-SNN). Moreover, it is a significant improvement that even reaches the best result achieved by ANNs on this dataset (92.4\% by CNN \citep{cramer_heidelberg_2020}). 
The SHD dataset contains rich temporal information, which challenges the model's capacity to extract temporal features \cite{cramer_heidelberg_2020}; hence, there is considerable effort required to develop SNN models using recurrent structures \cite{cramer_heidelberg_2020,yin_effective_2020,perez-nieves_neural_2021,yin_accurate_2021}.
Based on the recurrent structure, TA-SNN employs temporal-wise attention and a particular LIF neuron (LIAF \cite{wu_liaf-net_2021} that directly transmits membrane potential) to get an excellent result on SHD (91.08\%), outperforming LSTM (89\% \cite{cramer_heidelberg_2020}) but falling short of the result (92.4\%) produced by CNN processing (directly as 2D image input).
In contrast, instead of the recurrent layers, we use a simple fully-connected network with two hidden layers and successfully obtain the SOTA result by adding the proposed STSC module. 
For the first time, our model obtained CNN-like performance on the SHD dataset, which represents a substantial effort to illustrate the SNNs' potential.

\begin{figure}[t]
\centering
\includegraphics[width=0.9\columnwidth]{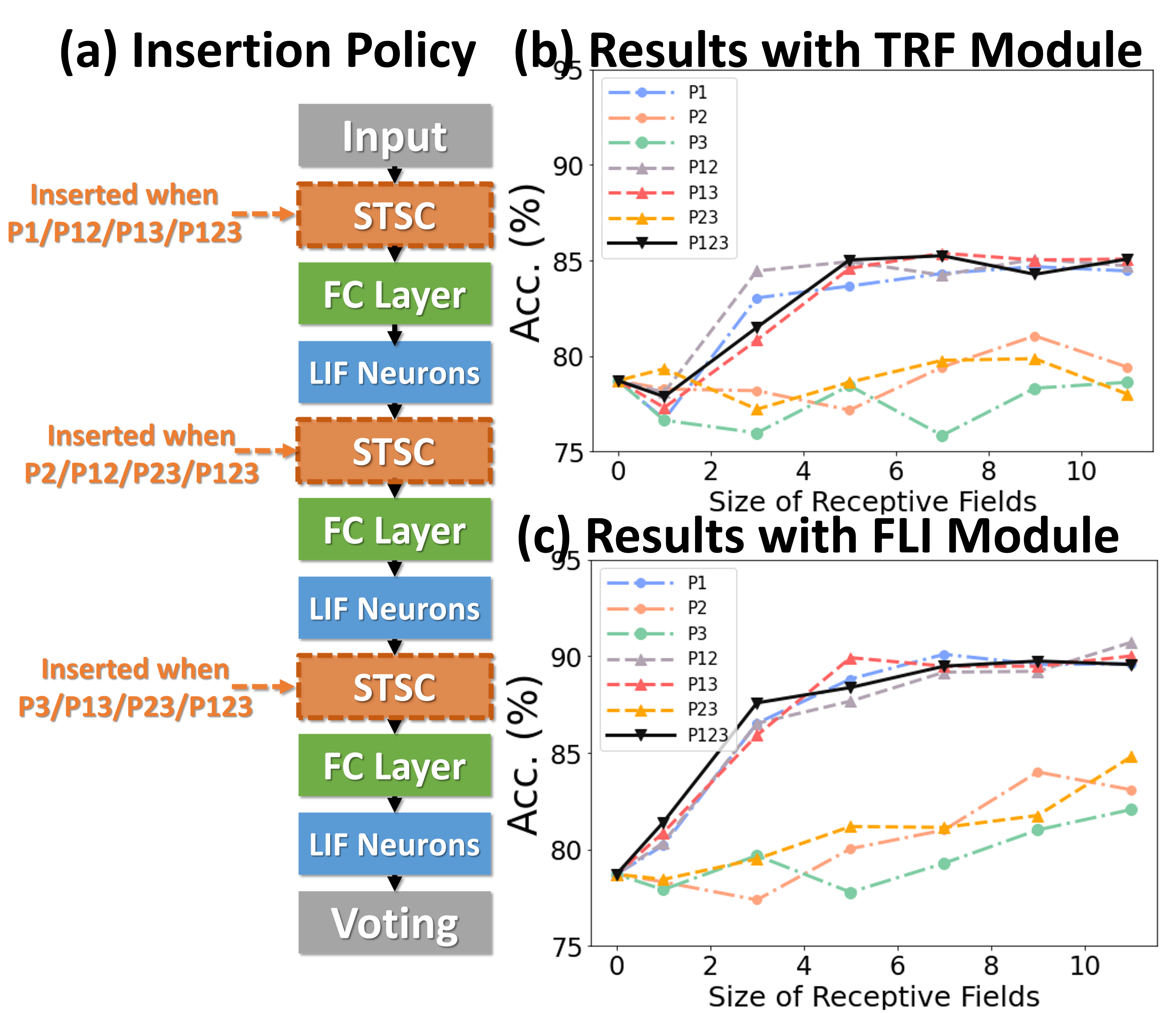} 
\caption{Control Experiments on SHD. (a) Policies of inserting STSC layers; (b) Accuracy comparison of TRF module vias different RFs; (c) Accuracy comparison of FLI module vias different RFs.
}.
\label{fig4}
\end{figure}

\begin{figure}[t]
\centering
\includegraphics[width=0.9\columnwidth]{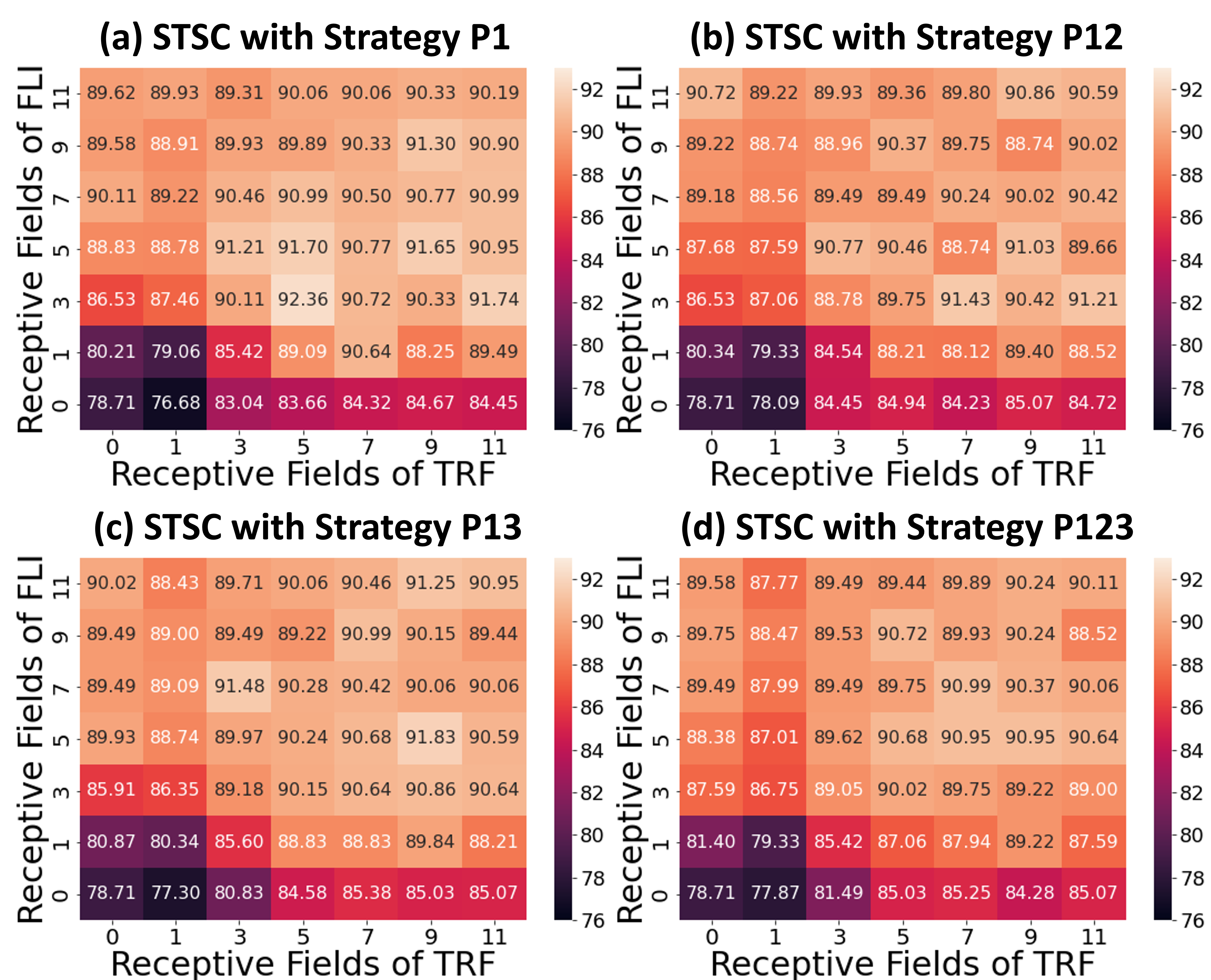} 
\caption{Influence of Receptive Fields on SHD. (a-d) show the experimental results vias RFs with P1/P12/P13/P123.
}
\label{fig5}
\end{figure}

\begin{figure}[t]
\centering
\includegraphics[width=0.9\columnwidth]{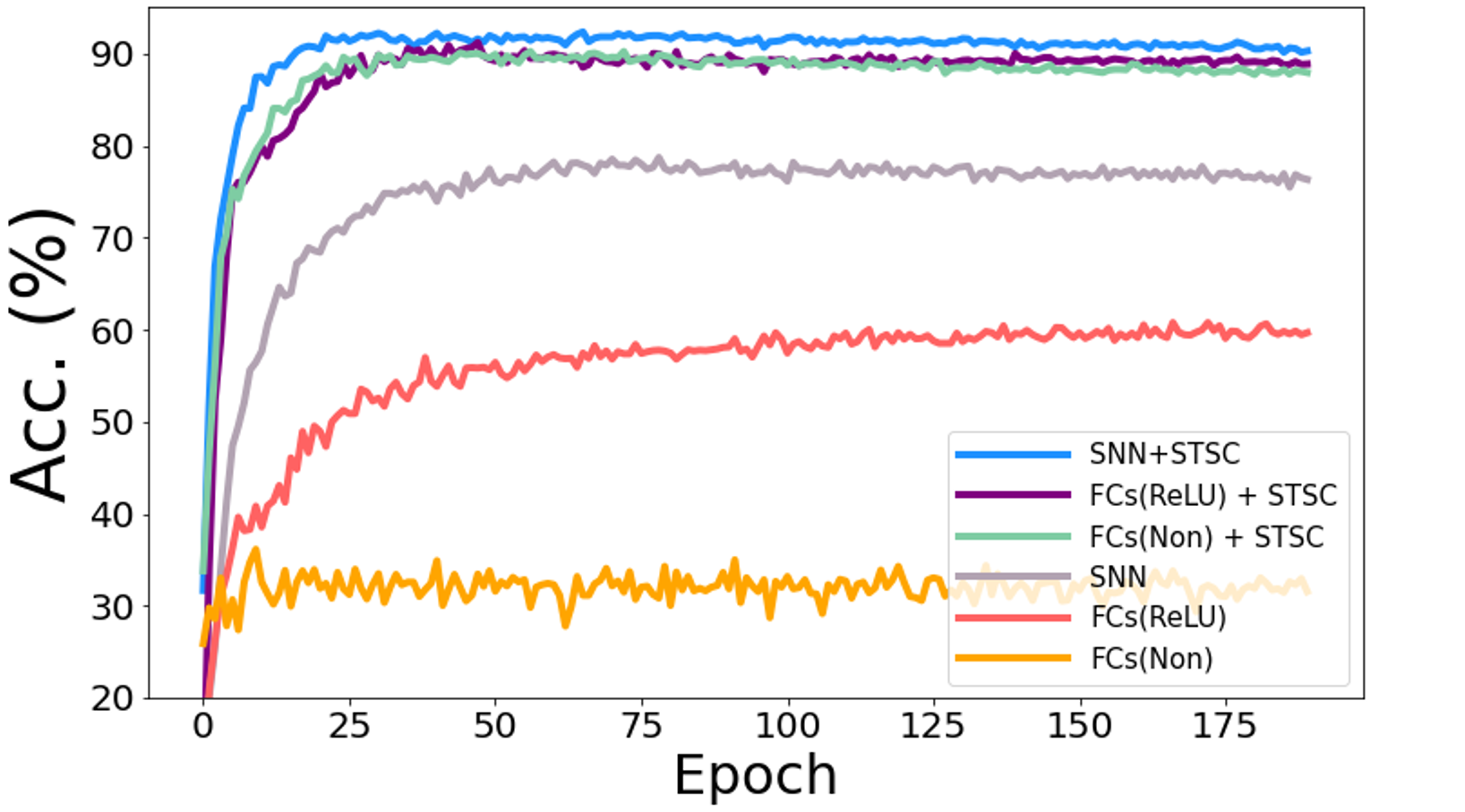} 
\caption{Ablation Study of Temporal Modules in SNNs. The accuracy comparison of different model vias training epochs on SHD.
'FCs(Non)' denotes the FC structure without LIFs and activation functions, 'FCs(ReLU)' denotes the FC structure with ReLU functions behind the first two FC layers, and 'SNN' denotes the FC structure with LIFs behind all three FC layers. Then, STSC modules are added just behind input (P1) in three models as a comparison.
}.
\label{fig6}
\end{figure}

\subsection{Control Experiments and Ablation Study}
To analyze the impact of each component on performance, we conduct control experiments on SHD. The SHD experiment is based on the fully-connected (FC) structure (see Tab. \ref{tab:3}),
with STSC modules strategically placed in front of the FC layers.
There are seven insertion strategies designated P1, P2, P3, P12, P13, P23, and P123 (see Fig. \ref{fig4}a).
Fig. \ref{fig4}b and \ref{fig4}c show the effect of varying receiving fields when TRF and FLI are used individually.
TRF reaches 85.38\% at P13 and RF=7, while FLI reaches 90.72\% at P12 and RF=11. 
Comparing the two modules reveals that the FLI module plays a major role in performance improvement; thus, it is crucial to offer a gating mechanism that introduces nonlinear expressions to FC layers.
Fig. \ref{fig4}b and \ref{fig4}c demonstrate that when the FLI or TRF modules are positioned in the first layer (P1/P12/P13/P123), they have a greater impact on performance than when they are positioned in the deep layer (P2/P3/P23). This suggests that the extraction of temporal features is more advantageous in shallow layers.
As shown in Fig. \ref{fig5}, we evaluated the impact of varying STSC  receptive fields on SHD performance.
Notably, raising the receptive field suitably will increase performance, whereas an overly broad receptive field setting would reduce accuracy.
We claim the performance drop is a result of the model's excessive expressive capacity, which overfits the train data.
This phenomenon is analogous to the usage of spatial 2D convolutions, in which the kernel size must be carefully determined.
Fig. \ref{fig5} indicates that the combination of TRF and FLI modules improves performance, demonstrating their complementarity.
Under the P1 strategy, setting TRF's RF=5 and FLI's RF=3 yields the best result of 92.36\%, with just one STSC added after the input layer.

\subsection{Analysis of Temporal Modules in SNNs}
In the vanilla SNNs, only neurons perform temporal operations; hence, its temporal feature extraction is predicated solely on the temporal dependence inside each neuron.
In order to assess the influence of temporal modules, we conduct the ablation study with LIF neurons and STSC modules on SHD datasets (see Fig. \ref{fig6}), based on the same FC structure (see Tab. \ref{tab:3}).
Experiments comparing 'FC(ReLU)' and 'SNN' demonstrate that utilizing LIF neurons to replace the activation function in the FC structure can definitely increase the performance of the SHD classification task, proving the LIF's capacity to handle temporal information and capture temporal features. 
Moreover, the 'FCs(Relu)+STSC' and 'FCs(non)+STSC' structures generated by adding the STSC module obtain greater performance than the vanilla 'SNN' model, demonstrating that our STSC module has superior temporal feature extraction capacity than LIF; hence, the utilization of time relationships within synaptic connections is valid and meaningful. 
Furthermore, integrating the STSC module and LIF concurrently inside the 'SNN+STSC' model achieves the highest performance, proving that time-dependent interactions in both synapses and neurons could coexist and be coordinated to perform better temporal information processing.
\section{Conclusion}
This work proposes to endow synaptic structures with spatio-temporal receptive fields and additional temporal dependencies in an effort to enhance the temporal information processing capabilities of SNNs. 
We propose the STSC module from the standpoints of both computational models and biological realities, which consists of TRF and FLI, implemented with temporal convolution and attention mechanisms. 
We verified the method's reliability on neuromorphic datasets of SHD, N-MNIST, CIFAR10-DVS, and DVS Gesture. 
Notably, the STSC supports SNNs in reaching the SOTA result (92.36\%) on the SHD dataset, which is comparable to ANNs' methods (89\% and 92.4\%), validating the potential of SNNs in the spatio-temporal data processing.

\bibliography{aaai23}

\end{document}